\documentclass[sigconf]{acmart}

\AtBeginDocument{%
  \providecommand\BibTeX{{%
    \normalfont B\kern-0.5em{\scshape i\kern-0.25em b}\kern-0.8em\TeX}}}

\setcopyright{acmcopyright}
\copyrightyear{2018}
\acmYear{2018}
\acmDOI{10.1145/1122445.1122456}

\acmConference[Woodstock '18]{Woodstock '18: ACM Symposium on Neural
  Gaze Detection}{June 03--05, 2018}{Woodstock, NY}
\acmBooktitle{Woodstock '18: ACM Symposium on Neural Gaze Detection,
  June 03--05, 2018, Woodstock, NY}
\acmPrice{15.00}
\acmISBN{978-1-4503-XXXX-X/18/06}

\makeatletter
\newcommand{\thickhline}{%
    \noalign {\ifnum 0=`}\fi \hrule height 1.5pt
    \futurelet \reserved@a \@xhline
}
\makeatother

\usepackage{multirow}
\usepackage{subfig}

\begin{document}
\def\x{{\mathbf x}}
\def\L{{\cal L}}
\def\eg{\textit{e.g.}}
\def\ie{\textit{i.e.}}
\def\Eg{\textit{E.g.}}
\def\etal{\textit{et al.}}
\def\etc{\textit{etc}}
\newcommand{\blue}[1]{\textcolor{blue}{#1}}

\title{\emph{Not made for each other}-- Audio-Visual Dissonance-based Deepfake Detection and Localization}

%
\author{Komal Chugh}
\affiliation{%
  \institution{Indian Institute of Technology Ropar}
  }
\email{2016csb1124@iitrpr.ac.in}
\author{Parul Gupta}
\affiliation{%
  \institution{Indian Institute of Technology Ropar}
  }
\email{2016csb1124@iitrpr.ac.in}

\author{Abhinav Dhall}
\affiliation{%
  \institution{Monash University/Indian Institute of Technology Ropar}
  }
\email{abhinav.dhall@monash.edu}
\author{Ramanathan Subramanian}
\affiliation{%
  \institution{Indian Institute of Technology Ropar}
  }
\email{s.ramanathan@iitrpr.ac.in}
%
%


\begin{abstract}
We propose detection of \emph{deepfake} videos based on the dissimilarity between the audio and visual modalities, termed as the \emph{\textbf{Modality Dissonance Score}} (MDS). We hypothesize that manipulation of either modality will lead to dis-harmony between the two modalities, \eg, loss of lip-sync, unnatural facial and lip movements,~\etc. MDS is computed as an aggregate of dissimilarity scores between audio and visual segments in a video. Discriminative features are learnt for the audio and visual channels in a chunk-wise manner, employing the \emph{cross-entropy} loss for individual modalities, and a \emph{contrastive} loss that models inter-modality similarity. Extensive experiments on the DFDC and DeepFake-TIMIT Datasets show that our approach outperforms the state-of-the-art by up to {7\%}. We also demonstrate temporal \emph{forgery localization}, and show how our technique identifies the manipulated video segments.
\end{abstract}




\maketitle

\section{Introduction}
\label{sec:introduction}
The combination of ubiquitous multimedia and high performance computing resources has inspired numerous efforts to \emph{manipulate} audio-visual content for both benign and sinister purposes. Recently, there has been a lot of interest in the creation and detection of high-quality videos containing facial and auditory manipulations, popularly known as \emph{deepfakes}~\cite{dolhansky2019deepfake,DBLP:DF-TIMIT}. Since fake videos are often indistinguishable from genuine counterparts, detection of deepfakes is challenging but topical given their potential for denigration and defamation, especially against women and in propagating misinformation~\cite{dystopia,AI2018}.  


Part of the challenge in detecting deepfakes via artifical intelligence (AI) approaches is that deepfakes are themselves created via AI techniques. Neural network-based architectures like Generative Adversarial Networks (GANs)~\cite{NIPS2014_5423} and Autoencoders~\cite{Vincent2008} are used for generating fake media content, and due to their `learnable' nature, output deepfakes become more naturalistic and adept at cheating fake detection methods over time. Improved deepfakes necessitate novel fake detection (FD) solutions; FD methods have primarily looked at frame-level visual features~\cite{Capsule_Forensics} for statistical inconsistencies, but  temporal characteristics~\cite{using_RNNs} have also been examined of late. Recently, researchers have induced audio-based manipulations to generate fake content, and therefore, corruptions can occur in both the visual and audio channels.

Deepfakes tend to be characterized by visual inconsistencies such as a lack of lip-sync, unnatural facial and lip appearance/movements or assymmetry between facial regions such as the left and right eyes (see~Fig.~\ref{fig:localisation} for an example). Such artifacts tend to capture user attention. Authors of~\cite{Grimes1991MildAD} performed a psycho-physical experiment to study the effect of \emph{dissonance}, \ie, lack of sync between the audio and visual channels on user attention and memory. Three different versions of four TV news stories were shown to users, one having perfect audio-visual sync, a second with some asynchrony, and a third with no sync. The study concluded that out-of-sync or \emph{dissonant} audio-visual channels induced a high user cognitive load, while in-sync audio and video (no dissonance condition) were perceived as a \emph{single stimulus} as they `belonged together'.

We adopt the \emph{dissonance} rationale for deepfake detection, and since a fake video would contain either an altered audio or visual channel, one can expect higher audio-visual dissonance for fake videos than for real ones. We measure the audio-visual dissonance in a video via the \emph{\textbf{Modality Dissonance Score}} (MDS), and use this metric to label it as \emph{real}/\emph{fake}. Specifically, audio-visual dissimilarity is computed over 1-second video chunks  to perform a fine-grained analysis, and then aggregated over all chunks for deriving MDS, employed as figure-of-merit for video labeling.

MDS is modeled based on the \emph{contrastive} loss, which has traditionally been employed for discovering lip-sync issues in video~\cite{outoftime}. Contrastive loss enforces the video and audio features to be \emph{closer} for real videos, and \emph{farther} for fake ones. Our method also works with videos involving only the audio/visual channel as our neural network architecture includes the video and audio-specific sub-networks, which seek to independently learn discriminative real/fake features via the imposition of a \textit{cross-entropy} loss. Experimental results confirm that these unimodal loss functions facilitate better real-fake discriminability over modeling only the contrastive loss. Experiments on the DFDC~\cite{dolhansky2019deepfake} and DF-TIMIT~\cite{VID-TIMIT} datasets show that our technique outperforms the state-of-the-art by upto 7\%, which we attribute to the following factors: (1) Modeling unimodal losses in addition to the contrastive loss which measures modality dissonance, and (2) Learning discriminative features by comparing 1-second audio-visual \emph{chunks}~to compute MDS, as against directly learning video-level features. Chunk-wise learning also enables \textit{localization} of transient video forgeries (\ie, where only some frames in a sequence are corrupted), while past works have only focused on the \emph{real}-\emph{fake} labeling problem. 

The {key contributions} of our work are: (a) We propose a novel multimodal framework for deepfake video detection based on modality dissonance computed over small temporal segments; (b) To our knowledge, our method is the first to achieve temporal forgery localization, and (c) Our method achieves state-of-art results on the DFDC dataset, improving AUC score by up to 7\%.

\section{Literature Review}
Recently, considerable research efforts have been devoted to detecting fake multimedia content automatically and efficiently. Most video faking methods focus on manipulating the video modality; audio-based manipulation is relatively rare. Techniques typically employed for generating fake visual content include 3D face modeling, computer graphics-based image rendering, GAN-based face image synthesis, image warping, \etc. Artifacts are synthesized either via face swapping while keeping the expressions intact (\eg, DeepFakes\footnote{https://github.com/deepfakes/faceswap}, FaceSwap\footnote{https://github.com/MarekKowalski/FaceSwap/}) or via facial expression transfer, \ie, facial reenactment (\eg, Face2Face \cite{thies2016face}). Approaches for fake video detection can be broadly divided into three categories as follows:

\subsection{Image-based} These methods employ image/frame-based statistical inconsistencies for \emph{real}/\emph{fake} classification. For example,~\cite{visual_artifacts} uses visual artifacts such as missing reflections and incomplete details in the eye and teeth regions, inconsistent illumination and heterogeneous eye colours as cues for fake detection. Similarly,~\cite{inconsistent_headposes} relies on the hypothesis that the 3-D head pose generated using the entire face's landmark points will be very different from the one computed using only the central facial landmarks, in case the face is synthetically generated. Authors of \cite{Li2018ExposingDVFWA}~hypothesize that fake videos contain artifacts due to resolution inconsistency between the warped face region (which is usually blurred) and surrounding context, and models the same via the VGG and ResNet deep network architectures. A capsule network architecture is proposed in~\cite{Capsule_Forensics} for detecting various kinds of spoofs, such as video replays with embedded images and computer-generated videos.

\begin{sloppypar}
A multi-task learning framework for fake detection-cum-segmentation of manipulated image (or video frame) regions is presented in~\cite{multi-task}. It is based on a convolutional neural network, comprising an encoder and a Y-shaped decoder, where information gained by one of the detection/segmentation tasks is shared with the other so as to benefit both tasks. A two-stream network is proposed in~\cite{Two-stream}, which leverages information from local noise residuals and camera characteristics. It employs a GoogLeNet-based architecture for one stream, and a patch based triplet network as second stream. Authors of~\cite{afchar2018mesonet} train a CNN, named MesoNet, to classify real and fake faces generated by the DeepFake and Face2face techniques. Given the similar nature of falsifications achieved by these methods, identical network structures are trained for both problems by focusing on mesoscopic properties (intermediate-level analysis) of images.  
\end{sloppypar}
    
\subsection{Video-based} 
Video-based fake detection methods also use temporal features for classification, since many a time, deepfake videos contain realistic frames but the warping used for manipulation is temporally inconsistent. For instance, variations in eye-blinking patterns are utilized in~\cite{in_ictu_oculi} to determine real and fake videos. Correlations between different pairs of facial action units across frames are employed for forgery detection in~\cite{protecting_leaders}. Authors of~\cite{using_RNNs} extract frame-level CNN features, and use them to train a recurrent neural network for manipulation detection.
    
\subsection{Audio-visual features based} 
Aforementioned approaches exploit only the visual modality for identifying deepfake videos. However, examining other modalities such as audio signal in addition to the video can also be helpful. As an example, authors of~\cite{mittal2020emotions} propose a Siamese network-based approach, which compares the multimedia as well as emotion-based differences in facial movements and speech patterns to learn differences between real and fake videos. Lip-sync detection in unconstrained settings is achieved by learning the \emph{closeness} between the audio and visual channels for in-sync vs out-of-sync videos via contrastive loss modeling in~\cite{outoftime}. While this technique is not meant to address deep-fake detection \emph{per se}, lip-sync issues can also be noted from a careful examination of deepfake videos, and the contrastive loss is useful for \emph{tying up} genuine audio-visual pairs. 

\subsection{Bimodal Approaches}
While it may be natural to see audio and video as the two main information modes that indicate the genuineness/fakeness of a video, one can also employ multiple cues from the visual modality for identifying fake videos. Multimodal cues are especially useful while tackling sophisticated visual manipulations. In~\cite{using_RNNs}, both intra-frame and inter-frame visual consistencies are modeled by feeding in frame-level CNN features to train a recurrent neural network. Statistical differences between the warped face area and the surrounding regions are learned via the VGG and ResNet architectures in~\cite{Li2018ExposingDVFWA}. Hierarchical relations in image (video frame) content are learned via a Capsule Network architecture in~\cite{Capsule_Forensics}.

\subsection{Analysis of Related Work}
Upon examining related literature, we make the following remarks to situate our work with respect to existing works.

\begin{figure*}[!tbph]
    \centering
    \includegraphics[width=\textwidth]{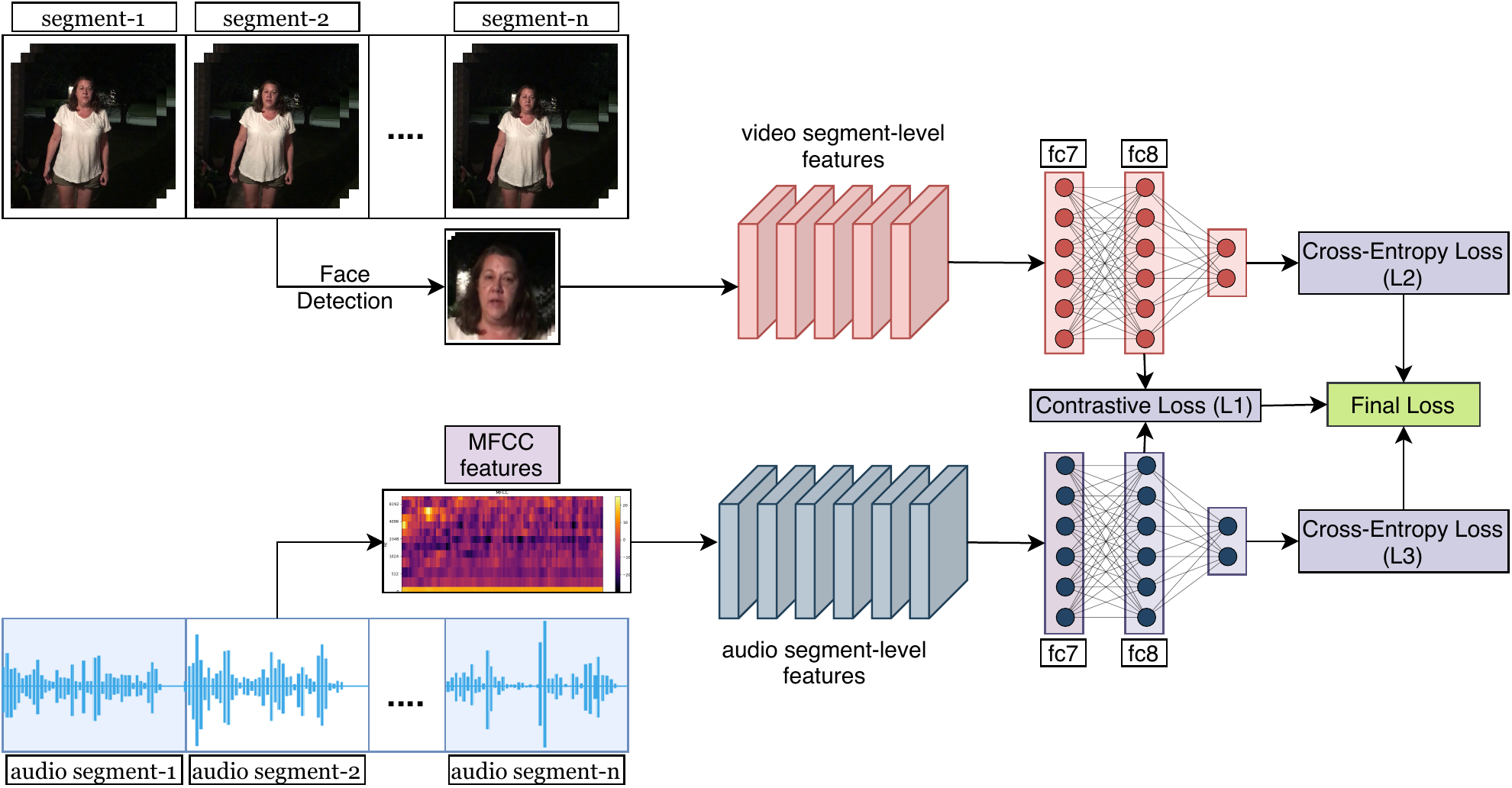}
    \caption{MDS-based fake video detection: Features extracted from 1-second audio-visual segments are input to the MDS network. The MDS network comprises the audio and visual sub-networks, whose description is provided in Table~\ref{tab:stream_arch}. Descriptors learned by the video and audio sub-networks are tuned via the cross-entropy loss, while the contrastive loss is employed to enforce higher dissimilarity between audio-visual chunks arising from \emph{fake} videos. MDS is computed as the aggregate audio-visual dissonance over the video length, and employed as a figure of merit for labeling a video as \emph{real}/\emph{fake}.}
    \label{fig:pipeline}
\end{figure*}

\begin{itemize}
\item[1.] While frame-based methods that learn spatial inconsistencies have been proposed to detect deepfakes, temporal-based approaches are conceivably more powerful to this end, as even if the manipulation looks natural in a static image, achieving temporally consistent warping even over a few seconds of video is difficult. Our approach models temporal characteristics, as we consider 1-second long audio-visual segments to distinguish between real and fake videos. Learning over tiny video chunks allows for a fine-grained examination of temporal differences, and also enables our method to temporally \emph{localize} manipulation in cases where the forgery targets only a small portion of the video. To our knowledge, prior works have only focused on assigning a \emph{real}/\emph{fake} video label.    
  
\item[2.] Very few approaches have addressed deepfake detection where the audio/video stream may be corrupted. In this regard, two works very related to ours are~\cite{outoftime} and~\cite{mittal2020emotions}. In~\cite{outoftime}, the contrastive loss function is utilized to enforce a smaller distance between lip-synced audio-visual counterparts; our work represents a novel application of the contrastive loss, employed for lip-sync detection in~\cite{outoftime}, to deepfake detection. Additionally, we show that learning audio-visual dissonance over small chunks and aggregating these measures over the video duration is more beneficial than directly learning video-level features. 

\item[3.] Both multimedia and emotional audio-visual features are learned for FD in~\cite{mittal2020emotions}. We differ from~\cite{mittal2020emotions} in three respects: (a) While we do not explicitly learn emotional audio-visual coherence, forged videos need not be emotional in nature; audio-visual consistency is enforced via the contrastive loss in our approach. (b) The training framework in~\cite{mittal2020emotions} requires a \emph{real}--\emph{fake} video pair. Our approach does not constrain the training process to involve video-pairs, and adopts the traditional training protocol. (c) Whilst~\cite{mittal2020emotions} perform a video-level comparison of audio-visual features to model dissonance, we compare smaller chunks and aggregate chunk-wise measures to obtain the MDS. This enables our approach to localize transient forgeries. 

\item[4.] Given that existing datasets primarily involve visual manipulations (number of datasets do not have an audio component), our architecture also includes audio and visual sub-networks which are designed to learn discriminative unimodal features via the cross-entropy loss. Our experiments show that additionally including the cross-entropy loss is more beneficial than employing only the contrastive loss. Enforcing the two losses in conjunction enables our approach to achieve state-of-the-art performance on the DFDC dataset.  

\end{itemize}

\section{MDS-based fake video detection}
As discussed in Section \ref{sec:introduction}, our FD technique is based on the hypothesis that deepfake videos have \emph{higher dissonance} between the audio and visual streams as compared to real videos. The dissimilarity between the audio and visual channels for a \emph{real}/\emph{fake} video is obtained via the Modality Dissonance Score (MDS), which is obtained as the aggregate dissimilarity computed over 1-second visual-audio chunks. In addition, our network enforces learning of discriminative visual and auditory features even if the contrastive loss is not part of the learning objective; this enables FD even if the input video is missing the audio/visual modality, in which case the contrastive loss is not computable. A description of our network architecture for MDS-based deepfake detection follows.  

%

\subsection{Overview}
Given an input video, our aim is to classify it as \emph{real} or \emph{fake}. We begin with a training dataset $D = \{(v^1,y^1), (v^2,y^2), ... , (v^N,y^N)\}$ consisting of $N$ videos. Here, $v^i$ denotes the input video and the label $y^i \ \epsilon \  \{0,1\}$ indicates whether the video is \emph{real} ($y^i=0$) or \emph{fake} ($y^i=1$).  The training procedure is depicted in Fig. \ref{fig:pipeline}. We extract the audio signal from input video $v^i$ using the \emph{ffmpeg} library, and split it into $D$-second segments. Likewise for the visual modality, we divide the input video into $D$-second long segments, and perform face tracking on these video segments using the S3FD face detector~\cite{zhang2017s3fd} to extract the face crops. This pre-processing gives us segments of visual frames $\{s^i_1, s^i_2, ... , s^i_n\}$ along with corresponding audio segments $\{a^i_1, a^i_2, ... , a^i_n\}$, where $n$ denotes segment count for an input video $v^i$.

We employ a bi-stream network, comprising the audio and video streams, for deepfake detection. Each video segment $s^i_t, t= 1 \ldots n$ is passed through a visual stream $S_v$, and the corresponding audio segment $a^i_t$ is passed through the audio stream $S_a$. These streams are described in Sections \ref{sec:visual} and \ref{sec:audio}. The network is trained using a combination of the \emph{contrastive loss} and the \emph{cross-entropy loss}. 
The contrastive loss is meant to \emph{tie up} the audio and visual streams; it ensures that the video and audio streams are \emph{closer} for real videos, and \emph{farther} for fake videos. Consequently, one can expect a low MDS for real, and high MDS for fake videos. If either the audio or visual stream is missing in the input video, in which case the contrastive loss is not computable, the video and audio streams will still learn discriminative features as constrained by the unimodal cross-entropy losses. These loss functions are described in Sec.~\ref{sec:loss}. 

\subsection{Visual Stream}
\label{sec:visual}


The input to the visual stream is $s^i_t$, a video sequence of size ($3 \times h \times w \times D*f$), where 3 refers to the RGB color channels of each frame, $h, w$ are the frame height and width, $D$ is the segment length in seconds, and $f$ is the video frame rate. Table \ref{tab:stream_arch} depicts the architecture of the video and audio sub-networks. The visual stream ($S_v$) architecture is inspired by the 3D-ResNet similar to \cite{DBLP:journals/corr/abs-1711-09577}. 3D-CNNs have been widely used for action recognition in videos, and ResNet is one of the most popular architectures for image classification. The feature representation learnt by the visual stream, in particular the fc8 output, denoted by $f_v$ is used to compute the contrastive loss. We also add a 2-neuron classification layer at the end of this stream, which outputs the visual-based prediction label. So the output of this stream, labeled as $\hat{y}_v^i$, constitutes the unimodal cross-entropy loss.




\definecolor{cadetblue}{rgb}{0.37, 0.62, 0.63}
\definecolor{pistachio}{rgb}{0.58, 0.77, 0.45}
\definecolor{orange}{rgb}{1.0, 0.5, 0.0}
\definecolor{palecopper}{rgb}{0.85, 0.54, 0.4}
\definecolor{mountainmeadow}{rgb}{0.19, 0.73, 0.56}
\definecolor{pastelorange}{rgb}{1.0, 0.7, 0.28}
\definecolor{yellow}{rgb}{1.0, 1.0, 0.0}
\definecolor{violet}{rgb}{0.56, 0.0, 1.0}

\begin{table}[t]
  \caption{Structure of the audio and visual streams (initial layers of the visual stream are the same as in the 3D ResNet architecture \cite{DBLP:journals/corr/abs-1711-09577}).}
  \label{tab:stream_arch}
  \scalebox{0.9}{
  \begin{tabular}{c c}
    \toprule
    \textbf{Visual Stream} & \textbf{Audio Stream}\\
    \midrule
    \multirow{3}{*}{\colorbox{red!30}{conv1}}  &   \colorbox{red!30}{conv\_1, 3$\times$3, 1, 64}\\
    
    & \colorbox{white}{batch\_norm\_1, 64}\\
    
     & \colorbox{blue!30}{pool\_1, 1$\times$1, MaxPool}\\ 
     
    \multirow{3}{*}{\colorbox{red!30}{conv2\_x}}   &  \colorbox{red!30}{conv\_2, 3$\times$3, 64, 192}\\
    
    & \colorbox{white}{batch\_norm\_2, 192}\\
    
     & \colorbox{blue!30}{pool\_2, 3$\times$3, MaxPool}\\  
    
    \multirow{2}{*}{\colorbox{red!30}{conv3\_x}}   &  \colorbox{red!30}{conv\_3, 3$\times$3, 192, 384}\\
    
    & \colorbox{white}{batch\_norm\_3, 384}\\

    \multirow{2}{*}{\colorbox{red!30}{conv4\_x}}   &  \colorbox{red!30}{conv\_4, 3$\times$3, 384, 256}\\ 
    
    & \colorbox{white}{batch\_norm\_4, 256}\\

    \multirow{3}{*}{\colorbox{red!30}{conv5\_x}}   &  \colorbox{red!30}{conv\_5, 3$\times$3, 256, 256}\\
    
    & \colorbox{white}{batch\_norm\_5, 256}\\
    
     & \colorbox{blue!30}{pool\_5, 3$\times$3, MaxPool}\\  
    
    \multirow{2}{*}{\colorbox{orange!80}{average pool}}  &  \colorbox{red!30}{conv\_6, 5$\times$4, 256, 512}\\ 
    
    & \colorbox{white}{batch\_norm\_6, 512}\\

    \colorbox{mountainmeadow!80}{fc7, 256$\times$7$\times$7, 4096} & \colorbox{mountainmeadow!80}{fc7, 512$\times$21, 4096} \\ 
    
    \colorbox{white}{batch\_norm\_7, 4096} & \colorbox{white}{batch\_norm\_7, 4096} \\

    \colorbox{mountainmeadow!80}{fc8, 4096, 1024} & \colorbox{mountainmeadow!80}{fc8, 4096, 1024} \\

    \colorbox{white}{batch\_norm\_8, 1024} & \colorbox{white}{batch\_norm\_8, 1024} \\

    \colorbox{cadetblue!80}{dropout, $p=0.5$}  & \colorbox{cadetblue!80}{dropout, $p=0.5$} \\

    \colorbox{mountainmeadow!80}{fc10, 1024, 2} & \colorbox{mountainmeadow!80}{fc10, 1024, 2} \\
    \bottomrule
  \end{tabular}}
\end{table}

\subsection{Audio Stream}
\label{sec:audio}
Mel-frequency cepstral coefficients (MFCC) features are input to the audio stream. MFCC features~\cite{Mogran2004} are widely used for speaker and speech recognition \cite{martinez2012speaker}, and have been the state-of-the-art for over three decades. For each audio segment of $D$ second duration $a^i_t$, MFCC values are computed and passed through the audio stream $S_a$. 13 mel frequency bands are used at each time step. Overall, audio is encoded as a heat-map image representing MFCC values for each time step and each mel frequency band.  We base the audio stream architecture on convolutional neural networks designed for image recognition. Contrastive loss $L_1$ uses the output of fc8, denoted by $f_a$. Similar to the visual stream, we add a classification layer at the end of the audio stream, and the output $\hat{y}_a^i$ is incorporated in the cross-entropy loss for the audio modality.

\begin{figure}[!b]
    \centering \vspace{-4mm}
    \subfloat{\includegraphics[width = 70mm]{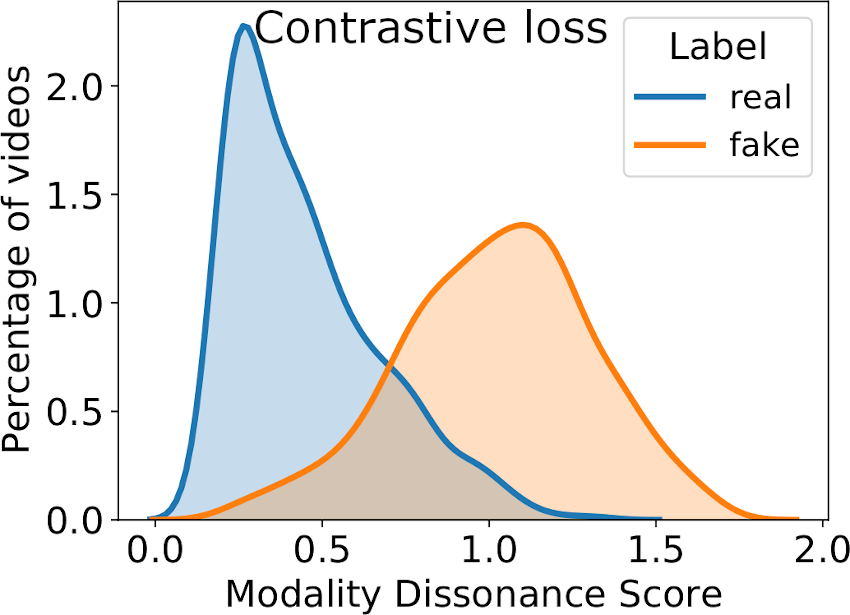}}
    \newline
    \subfloat{\includegraphics[width = 70mm]{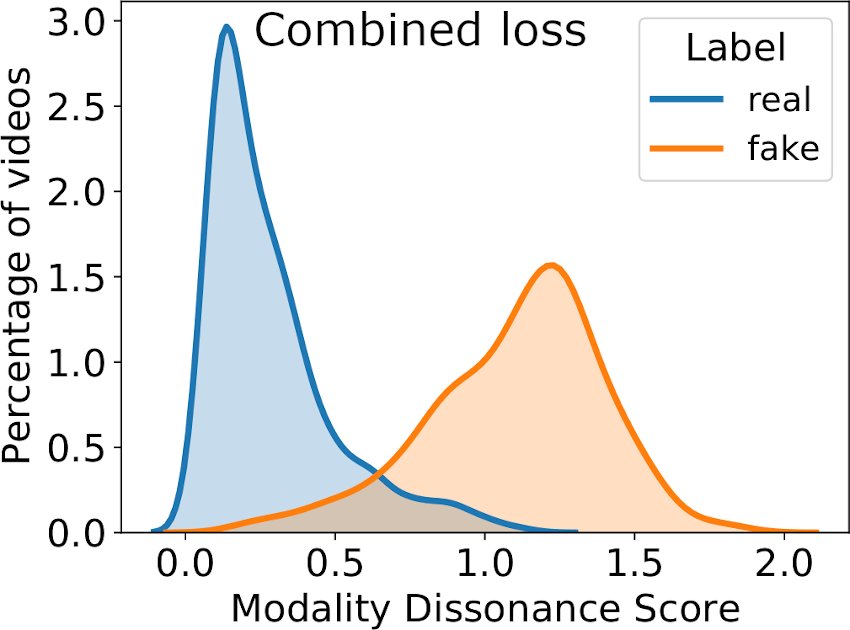}}
    \vspace{-1mm}
    \caption{Effect of loss functions: The graphs show the effect of audio and visual cross-entropy loss functions (Section \ref{sec:AblationStudies}). The top and bottom graphs show the MDS distribution for test samples, when the contrastive loss $ L_1$ alone and combined losses $L$ are used in the training of the MDS network, respectively.}\vspace{-1mm}
    \label{fig:Graphs}
\end{figure}

\subsection{Loss functions}
\label{sec:loss}
Inspired by~\cite{outoftime}, we use \emph{contrastive loss} as the key component of the objective function. Contrastive loss enables maximization of the dissimilarity score for manipulated video sequences, while minimizing the MDS for real videos. This consequently ensures separability of the real and fake videos based on MDS (see Fig.~\ref{fig:Graphs}). The loss function is represented by Equation \ref{eq:1}. Here, $y^i$ is the label for video $v^i$ and \emph{margin} is a hyper-parameter. Dissimilarity score $d_t^i$, is the Euclidean distance between the (segment-based) feature representations $f_v$ and $f_a$ of the visual and audio streams respectively. 

In addition, we employ the cross-entropy loss for the visual and audio streams to learn feature representations in a robust manner. These loss functions are defined in Equations \ref{eq:3} (visual) and \ref{eq:4} (audio). The overall loss $L$ is a weighted sum of these three losses, $L_1, L_2$ and $L_3$ as in Eq.~\ref{eq:9}.

\begin{equation}
    L_1 = \frac{1}{N} \sum_{i=1}^{N} (y^i) \ (d_t^i)^2 + (1-y^i) \ max(margin-d_t^i,0)^2 \label{eq:1} \\
\end{equation}

\begin{equation}
 d_t^i = \|f_v - f_a\|_2 \label{eq:2} 
\end{equation}

\begin{equation}
		L_2 = - \ \frac{1}{N} \sum_{i=1}^{N} y^i \ \log{\hat{y}_v^i} \ + (1-y^i) \ \log{(1-\hat{y}_v^i)} \label{eq:3} \\
\end{equation} 

\begin{equation}
	  L_3 = - \ \frac{1}{N} \sum_{i=1}^{N} y^i \ \log{\hat{y}_a^i} \ + (1-y^i) \ \log{(1-\hat{y}_a^i)} \label{eq:4}
\end{equation} 	

\begin{align} 
		 L =  \ \lambda_1L_1 + \lambda_2L_2 + \lambda_3L_3 \label{eq:9}
\end{align}
\noindent where $\lambda_1,\lambda_2, \lambda_3 =1$ in our design.

\subsection{Test Inference}
\label{sec:test}
During test inference, the visual segments $\{s^i_1, s^i_2, ... , s^i_n\}$ and corresponding audio segments $\{a^i_1, a^i_2, ... , a^i_n\}$ of a video are passed through $S_v$ and $S_a$, respectively. For each such segment, dissimilarity score $d_t^i$ (Equation \ref{eq:2}) is accumulated to compute the MDS as below: 

\begin{equation}
    MDS_i = \frac{1}{n} \sum_{t=1}^{n} d^i_t \label{eq:5}
\end{equation}

\noindent To label the test video, we compare $MDS_i$ with a threshold $\tau$ using $1\{MDS_i < \tau\}$ where $1\{.\}$ denotes the logical indicator function. $\tau$ is determined on the training set. We compute MDS for both real and fake videos of the train set, and the midpoint between the average values for the real and fake videos is used as a representative value for $\tau$.

\section{Experiments}
\subsection{Dataset Description}
As our method is multimodal, we use two public audio-visual deepfake datasets in our experiments. Their description is as follows:\\ \\
\textbf{Deepfake-TIMIT \cite{DBLP:DF-TIMIT}:} This dataset  contains videos of 16 similar looking pairs of people, which are manually selected from the publicly available VIDTIMIT \cite{VID-TIMIT} database and manipulated using an open-source GAN-based\footnote{https://github.com/shaoanlu/faceswap-GAN} approach. There are two different models for generating fake videos, one Low Quality (LQ), with $64\times64$ input/output size, and the other High Quality (HQ), with $128\times128$ input/output size. Each of the 32 subjects has 10 videos, resulting in a total of 640 face swapped videos in the dataset; each video is of $512\times384$ resolution with 25 fps frame rate, and of $\approx 4s$ duration. However, the audio channel is not manipulated in any of the videos.\\ 

\noindent \textbf{DFDC dataset \cite{dolhansky2019deepfake}:} The preview dataset, comprising of 5214 videos was released in Oct 2019 and the complete one with 119,146 videos in Dec 2019. Details of  the manipulations have not been disclosed, in order to represent the real adversarial space of facial manipulation. The manipulations can be present in either audio or visual or both of the channels.
    In order to bring out a fair comparison, we used 18,000 videos\footnote{Same as the videos used in \cite{mittal2020emotions}.} in our experiments. The videos are of $\approx10s$ duration each with an fps of 30, so there are $\approx300$ frames per video.

\subsection{Training Parameters}
For both the datasets, we used $D=1$ second segment duration and the margin hyper-parameter described in Equation \ref{eq:1} was set to 0.99. This resulted in (3 x 30 x 224 x 224) dimensional input for the visual stream in case of DFDC dataset and for the other dataset, Deepfake-TIMIT, the input dimension to the visual stream was (3 x 25 x 224 x 224). On DFDC we trained our model for 100 epochs with a batch size of 8 whereas for Deepfake-TIMIT the model only required 50 epochs with 16 batch size for convergence as the dataset size was small. We used Adam optimizer with a learning rate of 0.001 and all the results were generated on Nvidia Titan RTX GPU with 32 GB system RAM. For the evaluation, we use video-wise Area Under the Curve (AUC) metric.

\begin{table*}[!tbph]
  \caption{Comparison of our method with other techniques on DFDC and DFTIMIT datasets using the AUC metric. For our method frame-wise results are reported inside the brackets. (*We found a discrepancy in the train-test split of the 18k samples subset of \cite{mittal2020emotions}. Hence, the results have been updated after removing the discrepancy from the test set (20$^{th}$ March, 2021).)} 
  \label{tab:scores}
  \centering
  \scalebox{0.96}{
  \begin{tabular}{|c|c|c|c|c|c|c|c|c|c|}
    \cline{1-10}
    \multirow{2}{*}{}&
    \multicolumn{2}{ |c|  }{\multirow{2}{*}{} } &
    \multicolumn{7}{ |c| }{\textbf{Methods}}     \\ \cline{4-10}
    
      &      \multicolumn{1}{c}{}&           &
    Capsule & Multi-task  & HeadPose  & Two-stream  & VA-MLP  &  VA-LogReg  & Meso4   \\
    
    &      \multicolumn{1}{c}{}&           &
    \cite{Capsule_Forensics} & \cite{multi-task} & \cite{inconsistent_headposes} & \cite{Two-stream} & \cite{visual_artifacts} &  & \cite{afchar2018mesonet}   \\
    
    \cline{1-10}

    \multirow{3}{*}{\textbf{Datasets}} & \multicolumn{2}{|c|}{\textbf{DFDC}} & 
    53.3 & 53.6 & 55.9 & 61.4 & 61.9 &  66.2 &  75.3   \\ \cline{2-10}

     & \multirow{2}{*}{\textbf{DFTIMIT}} & \textbf{LQ} &
    78.4 & 62.2 & 55.1 & 83.5 & 61.4 &  77.0 &  87.8   \\ \cline{3-10}
    
      &   &  \textbf{HQ} &
    74.4 & 55.3 & 53.2 & 73.5 & 62.1 &  77.3 &  68.4  \\ 
  \cline{1-10}
    
    \textbf{Modality} & \multicolumn{2}{|c|}{} & 
    V & V & V & V & V &  V & V   \\ \cline{1-10}
 
    \thickhline
    \multirow{2}{*}{}&
    \multicolumn{2}{ |c|  }{\multirow{2}{*}{} } &
    \multicolumn{7}{ |c| }{\textbf{Methods}}     \\ \cline{4-10}
    
      &      \multicolumn{1}{c}{}&           &
    Xception-raw & Xception-c40  & Xception-c23  & FWA  & DSP-FWA  &  Siamese-based & \textbf{Our Method}    \\
    
    &      \multicolumn{1}{c}{}&           &
    \cite{rossler2019faceforensics++} &  &  & \cite{Li2018ExposingDVFWA} &  & \cite{mittal2020emotions} &     \\
    
    \cline{1-10}

    \multirow{3}{*}{\textbf{Datasets}} & \multicolumn{2}{|c|}{\textbf{DFDC}} & 
    49.9 & 69.7 & 72.2 & 72.7 & 75.5 &  \colorbox{red!30}{84.4} & \colorbox{blue!30}{\textbf{90.55 (90.66)*}}  \\ \cline{2-10}

     & \multirow{2}{*}{\textbf{DFTIMIT}} & \textbf{LQ} &
    56.7 & 75.8 & 95.9 & \colorbox{blue!30}{99.9} & \colorbox{blue!30}{99.9} &  96.3 & \colorbox{red!30}{97.9 (98.3)}   \\ \cline{3-10}
    
      &   &  \textbf{HQ} &
    54.0 & 70.5 & 94.4 & 93.2 & \colorbox{blue!30}{99.7} &  94.9 & \colorbox{red!30}{96.8 (94.7)}   \\ 
  \cline{1-10}
    
    \textbf{Modality} & \multicolumn{2}{|c|}{} & 
    V & V & V & V & V &  AV & AV  \\ \cline{1-10}
 
\end{tabular}}
\end{table*}

\begin{table}[b]
\caption{The AUC based comparison of audio stream, visual stream, contrastive loss only and combined loss  (Details in Section \ref{sec:AblationStudies}).} \label{tab:lambdas}
\scalebox{0.86}{
\begin{tabular}{|c|c|c|c|}
\hline
\begin{tabular}[c]{@{}c@{}}$\lambda_1$=0,$\lambda_2$=0,$\lambda_3$=1\\ Audio Stream\end{tabular} & \begin{tabular}[c]{@{}c@{}}$\lambda_1$=1,$\lambda_2$=0,$\lambda_3$=0\\ Contrastive Only\end{tabular} & \begin{tabular}[c]{@{}c@{}}$\lambda_1$=0,$\lambda_2$=1,$\lambda_3$=0\\ Visual Stream\end{tabular} & \begin{tabular}[c]{@{}c@{}}$\lambda_1$=1,$\lambda_2$=1,$\lambda_3$=1\\ Combined Loss\end{tabular} \\ \hline

  50.0                                                              & 86.1                                                                & 89.7        & 91.7                                                   \\ \hline
\end{tabular}}
\end{table}

\subsection{Ablation Studies} \label{sec:AblationStudies}
To decide the structure of the network and effect of different components, we chose 5800 videos (4000 real and 1800 fake) from the DFDC dataset, divided them into an 85:15 train-test split, following video-wise AUC and conducted the following experiments:\\
\textbf{Effect of Loss Functions:} We evaluated the contribution of the audio and visual streams based cross-entropy loss functions ($L_2$ and $L_3$, respectively). The hypothesis behind adding these two loss functions to the network is that the feature representations learnt across the audio and visual streams, respectively will be more discriminative towards the task of deep fake detection. This should further assist in the task of computing a segment dissimilarity score $d_t^i$, which disambiguates between fake and real videos. This hypothesis was tested by training the network on the DFDC dataset in two different settings. The first setting is based on training the MDS network with contrastive loss only. The second setting is combination of the three loss functions for training of the MDS network. Figure \ref{fig:Graphs} shows two graphs generated using the MDS scores as predicted by the networks from the two settings above. It is observed that the distributions of real and fake videos have lesser intersection in the case of the network trained with all three losses. Overall, combined loss function and contrastive loss only based networks achieved 92.7\% and 86.1\% AUC scores.

The difference attributes to the observation that the gap between average MDS for real and fake videos widens when cross-entropy loss functions are also used. Hence, there is more clear distinction between the dissonance scores for real and fake.  

\textbf{Audio and Visual Stream Performance:} We analysed the individual discriminative ability to identify fake and real videos for the audio and the visual streams. In this case the cross-entropy loss alone was used for training of the streams. It was observed that the audio stream only and visual stream only based deepfake classifiers achieved 50.0 and 89.7\%, respectively. Note that audio stream achieves less performance as in the DFDC dataset, minimal manipulation is performed on the audio signal of the videos.

\begin{figure}[t]
    \centering
    \includegraphics[width=4cm,height=3.5cm]{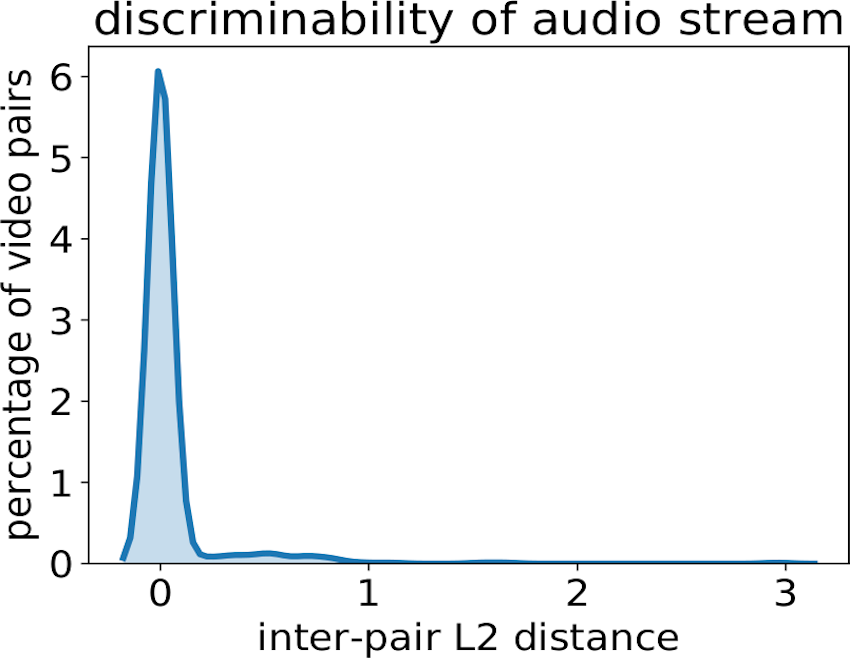}\hspace{2mm}
     \includegraphics[width=4.1cm,height=3.5cm]{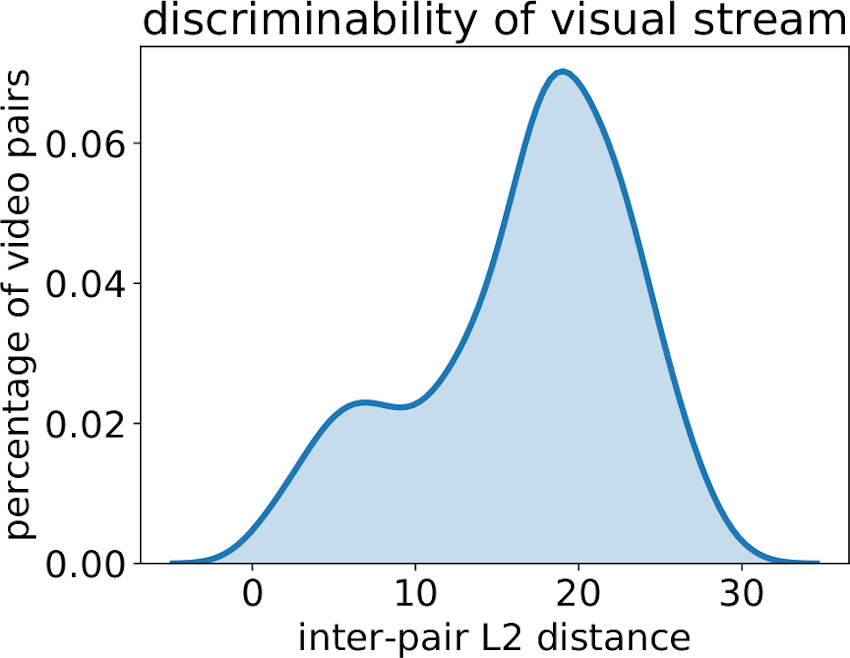}\vspace{-1mm}
    \caption{Discriminative-ability of individual streams: The graphs show the distribution of the mean L2 distance, when a fake segment and corresponding real segment are passed through only the audio and visual streams.}
    \label{fig:vid_aud_plot}
\end{figure}

\begin{figure}[b]
    \centering
        \vspace{-2mm}
    \includegraphics[width=8.6cm]{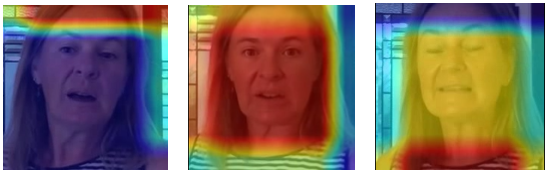}
    \caption{Grad-CAM results: Three frames are overlayed with the important attention regions highlighted using Grad-CAM. Note that forehead, face region and neck regions are highlighted in the 1st, 2nd and 3rd frames respectively.} \vspace{-4mm}
    \label{fig:GRADCAM}
\end{figure} 
In Equation \ref{eq:5}, for the four configurations: audio stream only, contrastive loss only, visual stream only and combined loss, we set the parameters as follows: ($\lambda_1=0,\lambda_2=0,\lambda_3=1$), ($\lambda_1=1,\lambda_2=0,\lambda_3=0$), ($\lambda_1=0,\lambda_2=1,\lambda_3=0$) and ($\lambda_1=1,\lambda_2=1,\lambda_3=1$). In the case of audio stream and visual stream based classification, the cross-entropy generated real and fake probabilities are generated segment-wise. We compute a maximum over the probabilities to compute the final fake and real for these two streams, individually. 

For further assessing the audio and visual stream's individual performances, we generated the plots shown in Figure \ref{fig:vid_aud_plot}. The process is as follows: a fake video segment and corresponding real video segment is passed through the streams and L2 distance is computed between the output of fc8 layer of the visual/audio sub-network. Then the average of these L2 distances for a video pair is plotted. It is observed in the audio stream plot, that most of the videos are centered close to 0 inter-pair L2 distance. This is due to the fact that audio has been modified in few cases in DFDC dataset. On the contrary, in the plot for the visual stream, we observed that the inter-pair L2 scores is spread across the dataset. This supports the argument that the visual stream is more discrimiantive due to the added cross entropy loss.  

In Table \ref{tab:lambdas}, we show the AUC of the four configurations mentioned above. Note that these numbers are on a smaller set of DFDC, which is used for ablation studies only. The contrastive loss only based configuration, which uses both the streams, achieves 86.1\%. The fusion of the cross-entropy loss into the final MDS network ($\lambda_1=1,\lambda_2=1,\lambda_3=1$ for Equation \ref{eq:5}), achieves 92.7\%. This backs up the argument that comparing (using contrastive loss) features learned through supervised channels enhances the performance of the MDS network.

\textbf{Segment Duration:} For deciding the optimal length of the temporal segment $D$, we conducted empirical analysis with temporal sizes of $D = [1,2,4]$ seconds. From the evaluation on the DFDC dataset, we observed that the temporal duration $D=1$ second is most optimum. This can attributed to the larger number of segments representing each video, thereby allowing fine-grained comparison of the audio and the visual data of a video.


\subsection{Evaluation on DFDC Dataset}
We compare the performance of our method on the DFDC dataset with other state-of-the-art works \cite{mittal2020emotions,Capsule_Forensics,afchar2018mesonet,rossler2019faceforensics++,visual_artifacts,in_ictu_oculi,inconsistent_headposes,Two-stream}. A total of 18,000 videos are used in this experiment\footnote{Please note that some earlier methods in Table \ref{tab:scores} are trained on DFDC preview (5000 videos), which is no longer available. }. In Table \ref{tab:scores}, it is observed that the MDS network approach outperforms the other visual-only and audio-visual based approaches by achieving 91.54\%. The performance is \textasciitilde8\% more relatively than the other audio-visual based approach \cite{mittal2020emotions}. Please note that the result 91.54\% is achieved with the test set containing 3281 videos out of which 281 videos have two subjects each. We chose the larger face and passed it through the visual stream. The performance of the network without these multiple subject videos is 93.50\%. We also report the frame-wise AUC (91.60\%) as mentioned in brackets in Table \ref{tab:scores}. This is computed by assigning each frame of a video the same label as predicted for the video by our method.

We argue that the gain in performance here is due to: (a) The temporal segmentation of the video into segment helps in fine-grained audio-visual feature comparison; (b) We extract task-tuned features from the audio and visual segments. Here task-tuned means that the features are learnt to discriminate between real and fake with the $L_2$ and $L_3$ loss functions, and (c) The visual stream's input is the face and an extra margin (see Figure \ref{fig:pipeline}) around it, which accounts for some rigid (head) movement along with the non-rigid (facial) movements. We visualise the important regions using the Gradient-weighted Class Activation Mapping (Grad-CAM) method \cite{selvaraju2017grad}. Figure \ref{fig:GRADCAM} shows the important regions localised by Grad-CAM on few frames of a video. Note that the region around the face is highlighted in the middle frame. Also, the forehead and the neck regions are highlighted in the first and third frames, respectively. This supports our argument that the disharmony between the non-rigid and rigid movement is also discriminative for the visual stream to classify between real and fake videos. 

\subsection{Evaluation on DFTIMIT Dataset}
The DeepFake-TIMIT (DFTIMIT) dataset is smaller as compared to the DFDC dataset. We trained the MDS network in two resolution settings: LQ and HQ. Table \ref{tab:scores} shows the comparison of our method with the other state-of-the-art methods. It is observed that our method achieves comparable results (LQ: 97.92\% and HQ: 96.87\%) with the top achieving method \cite{Li2018ExposingDVFWA}. In the DFTIMIT test set there are 96 videos in total. This applies that the penalty for mis-classification towards the overall AUC is high in DFTIMIT's case. It is interesting to note that our method mis-classified just 3 video samples in the HQ experiments and 2 videos in the LQ experiments. \cite{Li2018ExposingDVFWA} achieve state-of-the-art results (LQ: 99.9\% and HQ: 99.7\%) on DFTIMIT, however, achieve relatively lower AUC results (72.7\%) on the larger dataset DFDC. In comparison our method achieved 18\% more than \cite{Li2018ExposingDVFWA} on DFDC dataset. This could also mean that the DFTIMIT dataset is now saturated due to smaller size similar to the popular ImageNet dataset \cite{deng2009imagenet}. We also report the frame-wise AUC (LQ: 98.3\% and HQ: 94.7\%) for DFTIMIT  as mentioned in brackets in Table \ref{tab:scores}.

\begin{figure}[t]
    \centering
    \includegraphics[width=8.6cm]{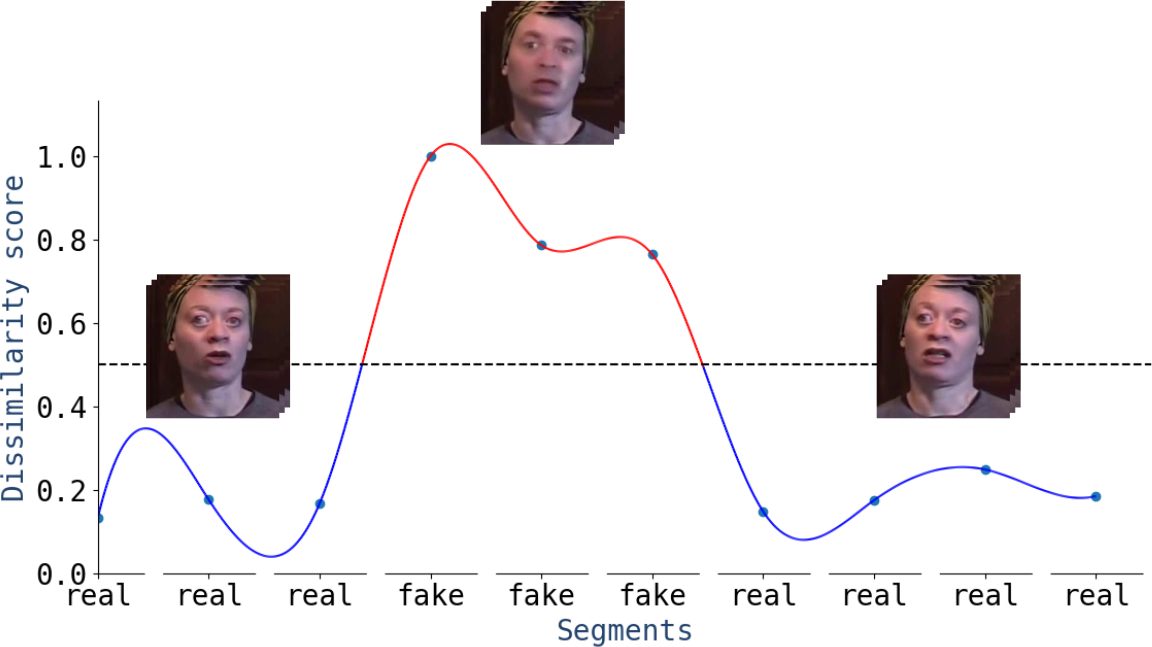} \vspace{6mm}
     \includegraphics[width=8.6cm]{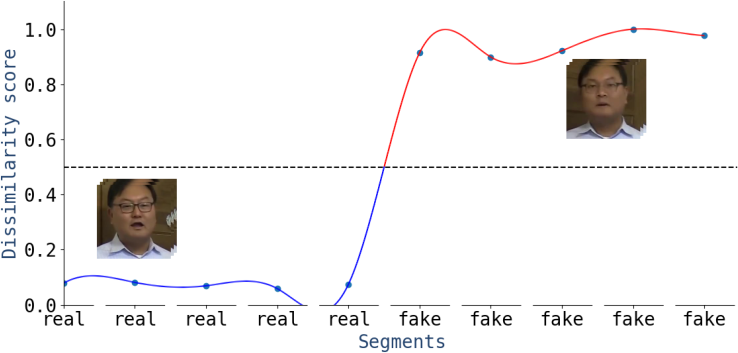}\vspace{-2mm}
    \caption{Forgery Localization results: In the two examples above, the red part of the curve is detected as fake and blue as real by our method. The segment of the video with segment wise dissimilarity above the threshold is labelled as fake and below the threshold is labelled as real. On the x-axis, we have the original segment labels and on the y-axis the segment-level dissimilarity score (Section \ref{sec:localisation}).}
    \label{fig:localisation}
\end{figure}

\subsection{Temporal Forgery Localization} \label{sec:localisation}
With the advent of sophisticated forgery techniques, it is possible that an entire video or smaller portions of the video are manipulated to deceive the audience. If in case parts of the original video are corrupted, it would be useful from the FD perspective to be able to locate the timestamps corresponding to the corrupted segments. In an interesting work, Wu \etal~\cite{wu2018busternet} proposed a CNN which detects forgery along with the forged locations in images. However, their method is only applicable to copy-paste image forgery. As we process the vidseo by dividing it into temporal segments, a fine-grained analysis of the input video is possible, thereby enabling \textit{forgery localization}. In contrast to the MDS network, earlier techniques \cite{mittal2020emotions,Li2018ExposingDVFWA} computed features over the entire video. We argue that if a forgery has been performed on a small segment of a video, the forgery signatures in that segment may get diluted due to pooling across the whole video. A similar phenomenon is also observed in prior work relating to pain detection and localization \cite{sikka2014classification}. As the pain event could be short and its location is not labeled, the authors divide the video into temporal segments for better pain detection. 

Most of the datasets, including the ones used in this paper have data manipulation performed on practically the entire video. To demonstrate the effectiveness of our method for forgery localization in fake videos, we joined segments from real and corresponding fake videos of the same subject at random locations. In Figure \ref{fig:localisation}, we show the outputs on two videos created by mixing video segments of the same subject from the DFDC dataset. Here, the segment-wise score is shown on the $y$-axis. The segments for which the score is above a threshold are assigned as being fake (red on the curve) and below the threshold (blue color on the curve) are assigned are real. In addition to Figs.~\ref{fig:GRADCAM} and~\ref{fig:Graphs}, forgery localization makes the working of the MDS-based fake detection framework more \emph{explainable} and \emph{interpretable}.


\section{Conclusion and Future Work}
We propose a novel bimodal deepfake detection approach based on the modality dissonance score (MDS), which captures the similarity between audio and visual streams for real and fake videos thereby facilitating separability. The MDS is modeled via the contrastive loss computed over segment-level audiovisual features, which constrains genuine audio-visual streams to be \textit{closer} than fake counterparts. Furthermore, cross-entropy loss is enforced on the unimodal streams to ensure that they independently learn discriminative features. Experiments show that (a) the MDS-based FD framework can achieve state-of-the-art performance on the DFDC dataset, and (b) the unimodal cross-entropy losses provide extra benefit on top of the contrastive loss to enhance FD performance. Explainability and interpretability of the proposed approach are demonstrated via audio-visual distance distributions obtained for \emph{real} and \emph{fake} videos, Grad-CAM outputs denoting attention regions of the MDS network, and forgery localization results. 

Future work would focus on (a) incorporating human assessments (acquired via EEG and eye-gaze sensing) in addition to content analysis adopted in this work; (b) exploring algorithms such as multiple instance learning for transient forgery detection, and (c) achieving real-time forgery detection (accomplished by online intrusions) given the promise of processing audio-visual information over 1-second segments.

\section{Acknowledgement}
We are grateful to all the brave frontline workers who are working hard during this difficult COVID19 situation.

\bibliographystyle{ACM-Reference-Format}
\bibliography{acm_mm2020}

\end{document}